# COMPARE: Clinical Optimization with Modular Planning and Assessment via RAG-Enhanced AI-OCT: Superior Decision Support for Percutaneous Coronary Intervention Compared to ChatGPT-5 and Junior Operators

Wei Fang[1†], Chiyao Wang[1†], Wenshuai Ma[1†], Hui Liu[1], Jianqiang Hu[1], Xiaona Niu[1], Yi Chu[1], Mingming Zhang[1], Jingxiao Yang[1], Dongwei Zhang[1], Zelin Li[1], Pengyun Liu[1], Jiawei Zheng[1], Pengke Zhang[1], Chaoshi Qin[1], Wangang Guo[1], Bin Wang[1], Yugang Xue[1], Wei Zhang[1], Zikuan Wang[1], Rui Zhu[2], Yihui Cao[2], Quanmao Lu[2], Rui Meng[2], Yan Li[*]

1. Department of Cardiology, Tangdu Hospital, The Fourth Military Medical University, Xi'an, Shaanxi, 710038, China.
2. Research Institute of Tsinghua University in Shenzhen, Shenzhen, Guangdong, 518057, China.

[†]Co-first Author
These authors contributed equally to this work

*Corresponding authors
Yan Li, Department of Cardiology, The Second Affiliated Hospital, Air Force Medical University, Xi'an, Shaanxi, 710038, China.
prefleeyan@163.com
Tel: 13892890227

**Abstract**

**Background** While intravascular imaging, particularly optical coherence tomography (OCT), improves percutaneous coronary intervention (PCI) outcomes, its interpretation is operator-dependent. General-purpose artificial intelligence (AI) shows promise but lacks domain-specific reliability. We evaluated the performance of CA-GPT-a novel large model deployed on an AI-OCT system-against that of the general-purpose ChatGPT-5 and junior physicians for OCT-guided PCI planning and assessment.

**Methods** In this single-center analysis of 96 patients (160 lesions) who underwent OCT-guided PCI, the procedural decisions generated by the CA-GPT, ChatGPT-5, and junior physicians were compared against an expert-derived procedural record. Agreement was assessed using ten pre-specified metrics across pre-PCI (5 metrics) and post-PCI (5 metrics) phases, generating a total score from 0 to 5.

**Results** For pre-PCI planning, CA-GPT demonstrated significantly higher median agreement scores (5[IQR 3.75-5]) compared to both ChatGPT-5 (3[2-4], $P<0.001$) and junior physicians (4[3-4], $P<0.001$). CA-GPT significantly outperformed ChatGPT-5 across all individual pre-PCI metrics and showed superior performance to junior physicians in stent diameter (90.3% vs. 72.2%, $P<0.05$) and length selection (80.6% vs. 52.8%, $P<0.01$). In post-PCI assessment, CA-GPT maintained excellent overall agreement (5[4.75-5]), which was significantly higher than both ChatGPT-5 (4[4-5], $P<0.001$) and junior physicians (5[4-5], $P<0.05$). Subgroup analysis confirmed CA-GPT's robust performance advantage in complex scenarios, including functionally significant lesions (OCT-FFR≤0.80) and acute coronary syndrome presentations.

**Conclusion** The CA-GPT-based AI-OCT system achieved superior decision-making agreement versus a general-purpose large language model and junior physicians across both PCI planning and assessment phases. By integrating automated OCT analysis with a curated knowledge base, this approach provides a standardized and reliable method for intravascular imaging interpretation, demonstrating significant potential to augment operator expertise and optimize OCT-guided PCI, particularly in complex scenarios.

**Keywords** CA-GPT-based AI-OCT, Optical coherence tomography, Percutaneous coronary intervention, Domain-specific artificial Intelligence, Large language models, Clinical decision-making

## Introduction

Cardiovascular disease (CVD) remains the leading cause of mortality and disability-adjusted life years (DALYs) globally, accounting for approximately 19.2 million deaths and 437 million DALYs annually, with coronary artery disease (CAD) as the major contributor.[1] Percutaneous coronary intervention (PCI) is performed in more than 4 million patients worldwide each year, representing a cornerstone of revascularization therapy.[2] Evidence has established that imaging-guided PCI, particularly using optical coherence tomography (OCT), can effectively reduce cardiovascular risks by optimizing stent implantation.[3-5] However, the full potential of OCT is often limited by its significant dependence on operator interpretation. Substantial Inter-observer variability in classifying calcium severity and thin-capped fibroatheroma (TCFA) has been reported, with intraclass correlation coefficients (ICC) as low as 0.55-0.61, particularly among less experienced physicians.[6]

In clinical practice-particularly in regions with uneven healthcare resource distribution-OCT interpretation skills may vary considerably.[7] Due to disparities in economic development, limited availability of structured imaging training, and high clinical workload, many interventional cardiologists in China face a slower learning curve in OCT interpretation.[8-11] This skill gap may lead to inconsistent treatment quality and prevent some patients from receiving optimal lesion assessment and stent deployment.[12] Addressing this gap requires new tools capable of assisting physicians in real-time OCT interpretation and PCI decision-making.

Artificial intelligence (AI) has emerged as a promising adjunct in cardiovascular imaging[13,14]. Among AI technologies, large language models (LLMs) such as ChatGPT-5 represent general-purpose systems trained on internet-scale corpora. However, their clinical applicability remains questionable. Published studies have shown that generative pre-trained transformer (GPT) models can achieve 60-85% accuracy on standardized medical licensing examinations but only 56% on general medical queries, which suggests that while ChatGPT shows promise, its current capabilities remain inadequate for standalone clinical decision-making.[15-17] This limitation is particularly pronounced in specialized domains like imaging-guided clinical tasks, where LLMs remain prone to hallucinations and factually unsupported outputs.[14,17] Retrieval-augmented generation (RAG) has been introduced as a strategy to mitigate hallucination by grounding model outputs in validated evidence sources. Integrating LLMs with RAG and structured imaging analysis therefore offers an opportunity to enhance accuracy and standardization in OCT interpretation.[18-20]

With recent progress in domain-specific cardiovascular models,[21-23] we developed an integrated AI-OCT system powered by the CA-GPT large model. This system is designed to provide end-to-end procedural support for PCI, where the CA-GPT functions as the central decision engine, nesting automated OCT image analysis within a RAG framework for clinical reasoning. To validate its clinical utility, we conducted a retrospective study evaluating the diagnostic agreement of the CA-GPT, a general LLM, and junior interventionalists compared with expert operators in OCT-guided PCI.

**Methods**

**Study Design and Population**
This analysis was conducted at Tangdu Hospital of Fourth Military Medical University. We included consecutive patients who underwent OCT-guided PCI between June 2024 and August 2025. Eligible cases were required to meet follow criteria: (i) Age between 18 and 85 years; (ii) complete OCT imaging of the target lesion, with adequate blood clearance and minimal motion artifacts to ensure valid analysis; and (iii) comprehensive procedural records documenting lesion assessment, pretreatment device type and size, stent diameter and length, and post-PCI optimization strategies. Exclusion criteria included chronic total occlusion lesions, bifurcation lesions requiring two-stent techniques, inability to advance the OCT catheter, thrombus burden precluding imaging, poor-quality OCT images, and incomplete procedural records lacking critical information. The study was approved by the medical ethics committee and was conducted in accordance with the the Declaration of Helsinki and applicable guidelines.[24] Informed consent was waived for this analysis of anonymized clinical data.

**Decision-Making Systems and Comparators**
The AI-OCT system was developed by integrating the CA-GPT large model (developed by the Research Institute of Tsinghua University, Shenzhen) with the OCT hardware (P80) provided by Vivolight Medtech (China). This system drives AI-powered OCT analysis and constitutes a domain-specific platform that integrates a "small- and large-model" framework to provide end-to-end procedural support for PCI. The small model layer conducts structured OCT analysis, performing 13 core functions-including lumen segmentation, plaque characterization, stent apposition assessment, and OCT-fractional flow reserve (FFR) computation-six of which are proprietary algorithms. The large model layer is built upon the open-source DeepSeek-R1 (a 14B parameter model released in November 2023, with a knowledge cutoff of October 2023; inference temperature=0.7, repetition penalty=1.1). The system employs RAG, a technique that allows LLMs to access guideline content during response generation, improving accuracy when answering questions based on the guidelines.[25] RAG integrates analytical outputs with a knowledge base of current guidelines and over 100,000 annotated PCI cases. This integration facilitates a "retrieve-reason-generate" cycle, producing structured, evidence-traceable recommendations for all PCI stages (**Figure 1**).

ChatGPT-5 (developed by OpenAI and officially released on August 7, 2025) served as a general-purpose AI comparator. For each case, key OCT parameters were manually extracted and entered into the model with prompts restricting its role to that of a "senior interventional cardiologist." The model was required to generate concise (<40 words) recommendations covering revascularization, pretreatment, stent selection, and optimization (**Figure S1**). Junior physicians were interventional cardiologists with 1-5 years of PCI experience who independently reviewed complete OCT images and provided decisions without access to expert or AI outputs. The reference standard was based on the actual procedural records from high-volume senior operators and was adjudicated by two independent senior experts (≥10 years of PCI experience, ≥200 annual cases) using predefined scoring criteria (**Table S1**).[26,27] Each decision was independently scored by both experts, with any discrepancies resolved by a third senior expert to achieve final consensus.

**Data Processing and Input**
Data processing followed a standardized workflow. For the CA-GPT-based AI-OCT system, raw OCT images were directly uploaded, allowing the CA-GPT large model to automatically drive the analysis pipeline for generation of quantitative parameters and recommendations. For ChatGPT-5, parameters were manually extracted for input, including pre-PCI (minimum lumen area [MLA], OCT-FFR, maximum index of plaque attenuation within 4 mm [Max IPA$_{4mm}$], calcium score, calcium volume, distal and proximal reference diameters, lesion length) and post-PCI (minimum stent area [MSA], stent expansion, apposition, dissections, tissue prolapse) variables. Junior physicians independently interpreted complete OCT pullbacks and formulated their own strategies. All outputs were subsequently compared against the actual procedural records.

**Evaluation criteria**
Ten predefined decision metrics were used to quantify agreement across pre-PCI and post-PCI phases: (1) revascularization judgment; (2) pretreatment strategy; (3) pretreatment device size (±0.5 mm tolerance); (4) stent diameter (±0.5 mm tolerance); (5) stent length (±5 mm tolerance); (6) MSA adequacy (threshold 4.5 mm²); (7) stent expansion (≤80% considered underexpanded); (8) stent apposition (severe malapposition defined as >400 μm or ≥1mm length); (9) severe edge dissection (arc ≥60° and length≥3mm); and (10) severe tissue prolapse (≥10% of stent area) (**Table S1**). The decision metrics were based on the Chinese Expert Consensus on the Application of OCT in Coronary Artery Disease and the Expert Consensus on Intravascular Imaging in PCI.[26,27] Given that not all patients underwent both pre- and post-PCI OCT imaging, the ten predefined decision metrics were separately scored for the pre-PCI (5 metrics) and post-PCI (5 metrics) phases. Each metric was assigned 1 point if consistent with the actual procedural records and 0 otherwise, yielding a total agreement score ranging from 0 to 5. The primary endpoint was this total agreement score for each comparator against the actual procedural records. Secondary endpoints included agreement on individual metrics and the identification of influencing factors.

**Statistical Analysis**
Statistical analyses were performed using R version 4.5.1 (R Foundation for Statistical Computing, Vienna, Austria). Continuous variables were expressed as mean ± standard deviation or median (interquartile range), and categorical variables were reported as counts (n) with percentages. The normality of continuous variables was assessed using the Shapiro-Wilk test. The total agreement scores (range: 0-5) among CA-GPT, ChatGPT-5, and junior physicians were compared using repeated measures ANOVA (parametric) or the Friedman test (non-parametric), with post-hoc pairwise comparisons conducted using paired t-tests or Wilcoxon signed-rank tests, as appropriate. For each individual question, the compliance rates were compared using Cochran's Q test, followed by pairwise McNemar tests. The Bonferroni method was applied to control the family-wise error rate at $\alpha = 0.05$ for all multiple comparisons. Subgroup analyses were performed according to vessel location (left anterior descending artery [LAD] vs. left circumflex/right coronary artery [LCx/RCA]), pre-PCI OCT-FFR (≤0.80 vs. >0.80), clinical presentation (ACS vs. SCAD) and calcium severity (Calcium score<4 vs. =4). Two-sided P values <0.05 were considered statistically significant.

## Results

### Study Population and Baseline Characteristics

From a consecutive cohort of 156 patients who underwent OCT-guided PCI, a total of 96 patients with 160 lesions met the study criteria and were included in the final analysis (**Figure S2**). The baseline clinical and lesion characteristics are summarized in **Table 1**. The study population had a mean age of 60.2±10.9 years, with 68.8% being male. Acute coronary syndromes (ACS) accounted for 82.8% of presentations, with unstable angina being the most common diagnosis (60.2%). The final PCI strategies comprised DES implantation (80.2%), DCB angioplasty (13.5%), or no immediate PCI (6.3%).

The OCT imaging analysis showed that the lesions were predominantly located in the LAD (60.6%), with a mean lesion length of 33.02±13.67 mm (**Table 1**). The mean MLA was 3.94±2.42 mm², indicating significant anatomical severity, which was corroborated by a functionally impaired mean pre-PCI OCT-FFR of 0.74±0.08. The majority of lesions (87.5%) exhibited mild calcification (Calcium score<4). Post-PCI, the mean MSA was 5.54±1.88 mm² with a mean stent expansion of 70.0±14.34%, and the functional significance of the stented lesion was resolved, as evidenced by a significantly improved mean post-PCI OCT-FFR of 0.89±0.06. The operator characteristics are detailed in **Table S2**. The PCI strategies for comparison were generated by four junior physicians (all with ≤5 years of experience), while the reference standard was derived from procedures performed by a separate group of senior operators, the majority of whom (85.7%) had ≥10 years of PCI experience and held senior academic titles.

### CA-GPT Demonstrates Superior Decision Agreement in Pre-PCI Planning

In pre-PCI decision-making, CA-GPT demonstrated significantly higher overall agreement scores compared to both ChatGPT-5 and junior physicians (median: 5[IQR 3.75-5] vs. 3[2-4] vs. 4[3-4], respectively; P<0.001) for all pairwise comparisons (**Table 2**, **Figure 2**). This superiority was consistent across most individual metrics (**Table 2**, **Figure S3**). Specifically, CA-GPT significantly outperformed ChatGPT-5 in pretreatment device type selection (73.6% vs. 37.5%, P<0.001), device sizing (70.8% vs. 40.3%, P<0.001), stent diameter selection (90.3% vs. 63.9%, P<0.001), and stent length selection (80.6% vs. 54.2%, P<0.001). Compared to junior physicians, CA-GPT showed superior performance in stent diameter (90.3% vs. 72.2%, P<0.05) and length selection (80.6% vs. 52.8%, P<0.01).

**CA-GPT Outperforms Comparators in Post-Procedural Assessment**
In post-PCI assessment, the overall agreement of CA-GPT (median score: 5 [4.75-5]) was significantly higher than that of ChatGPT-5 (4[4-5], P<0.001) and junior physicians (5[4-5], P<0.05) (**Table 2, Figure 2, Figure S3**). Analysis of individual endpoints revealed a distinct performance profile. All three groups achieved high agreement on MSA assessment (CA-GPT: 100%, ChatGPT-5: 100%, Junior physicians: 95.5%). For stent expansion, the agreement of CA-GPT (78.4%) was significantly higher than that of ChatGPT-5 (33.0%, P<0.001) and comparable to junior physicians (84.1%, P=1.000). In stent apposition assessment, CA-GPT (93.2%) performed significantly better than junior physicians (76.1%, P<0.05), while no significant difference was observed compared to ChatGPT-5 (88.6%, P=0.402).

**Subgroup Analysis Reveals Differential Advantages of CA-GPT**
Subgroup analyses delineated the specific scenarios where CA-GPT demonstrated its strongest utility (**Table S3**). In pre-PCI planning, CA-GPT's superiority over ChatGPT-5 was consistent across all subgroups (all P<0.05). Compared to junior physicians, its performance was context-dependent: CA-GPT achieved significantly higher agreement in LCx/RCA lesions (5[4-5] vs. 4[2-4], P<0.01), lesions with significant ischemia (OCT-FFR≤0.80; 5[4-5] vs. 4[3-4], P<0.001), ACS presentations (5[3.25-5] vs. 4[3-4.75], P<0.01), and mildly calcified lesions (5[4-5] vs. 4[3-4], P<0.001). However, no statistically significant difference was found between CA-GPT and junior physicians in LAD lesions (P=0.076), non-ischemic lesions (OCT-FFR >0.80; P=1.000), SCAD (P=1.000), or severely calcified lesions (P=0.708). In post-PCI assessment, the performance pattern shifted. CA-GPT maintained significant superiority over ChatGPT-5 in all subgroups. Its advantage over junior physicians was more localized, reaching statistical significance only in LCx/RCA lesions (5[5-5] vs. 5[4-5], P<0.05), while showing comparable performance in all other subgroups including LAD lesions (P=0.408), ACS (P=0.060), and SCAD (P=0.267).

**Representative Case Demonstrating High Decision Concordance of CA-GPT**
A representative case (**Figure 3**) demonstrates CA-GPT integrated decision-support capability. In a complex LAD lesion with significant calcification, the system provided comprehensive procedural planning including precise device sizing and optimization strategies, achieving perfect agreement with the actual expert-performed procedure across all evaluation metrics.

**Discussion**

This study presents a novel CA-GPT-based AI-OCT system that integrates intravascular imaging with a RAG framework for OCT-guided PCI decision support. Our comparative analysis reveals a distinct performance hierarchy: the RAG-enhanced CA-GPT-based system demonstrated superior agreement with expert consensus, outperforming both junior interventionalists and the general-purpose ChatGPT-5. This gradient not only validates the technical advantage of a purpose-built, integrated AI system in specialized medical domains but also highlights the transformative potential of integrating structured clinical knowledge with real-time image analysis.

In our study, the CA-GPT showed remarkable improvements in decision-making consistency across multiple PCI decision metrics. Specifically, stent diameter selection was accurately chosen 90.3% of the time by CA-GPT, outperforming junior physicians (72.2%, $P<0.05$) and ChatGPT-5 (63.9%, $P<0.001$). This consistent performance underscores the system's potential to mitigate the variability often introduced by operator-dependent factors, particularly for junior physicians with limited experience. Furthermore, the performance of CA-GPT in stent length selection (80.6% vs. 52.8%, $P<0.01$) was significantly higher than that of junior physicians, again demonstrating its superior utility in providing accurate and standardized PCI decisions.

The critical advantage of the CA-GPT-based AI-OCT system lies in its RAG architecture, which effectively bridges the gap between data-driven AI and evidence-based clinical practice. In contemporary interventional cardiology, the rapid evolution of guidelines and the overwhelming clinical workload often create challenges for physicians, particularly junior operators, in maintaining updated knowledge and consistent decision-making. The integrated AI-OCT system addresses this fundamental need by grounding its reasoning in a continuously updated knowledge base encompassing current guidelines and extensive real cases.[18-20] This approach ensures that each recommendation is not merely data-driven but is substantiated by validated clinical evidence, thereby creating a "computerized decision support system" that aligns with the core principles of evidence-based medicine.

The RAG framework proved crucial in mitigating the well-documented risks of AI "hallucinations" in clinical contexts.[28] General-purpose large language models, like ChatGPT-5, may produce outputs that sound plausible but lack factual accuracy when applied to specialized medical tasks. These "hallucinations" are especially problematic in high-stakes settings like PCI, where clinical decision-making is heavily dependent on precision. RAG addresses this issue by grounding the model's reasoning in a continuously updated knowledge base, ensuring that its outputs are not only data-driven but also substantiated by validated clinical evidence. Studies have shown that incorporating RAG enhances model accuracy by up to 18% in clinical reasoning tasks, particularly in complex scenarios such as diagnosing critical illness, where accurate, evidence-based decisions are critical for patient outcomes.[29-31] In our study, the CA-GPT-based system, by embedding domain-specific knowledge through RAG, significantly outperformed general-purpose models like ChatGPT-5 in PCI decision-making. This underscores the transformative potential of integrating RAG within a dedicated AI-OCT platform in improving clinical AI reliability and reducing errors in decision support systems.

While ChatGPT-5 demonstrated impressive linguistic capabilities, its performance in specialized medical tasks was hindered by a lack of tailored training and evidence-grounding. Several factors specific to our study context may have contributed to its suboptimal performance. First, inaccuracies in translating complex Chinese medical terminology into English for querying ChatGPT-5 could have introduced errors or caused loss of nuance. This is particularly important in fields like PCI, where precise terminology and context are crucial. Second, and more critically, ChatGPT-5's training data predominantly comes from Western populations and international guidelines (e.g., American or European), which may not fully align with the Chinese expert consensus that served as our reference standard.[26,27] Discrepancies in lesion preparation philosophy, stent sizing practices, or expansion targets between guidelines could directly explain some of the observed disagreements. Our findings demonstrated that ChatGPT-5 achieved significantly lower agreement rates across most PCI decision metrics compared to the CA-GPT-based system, highlighting the limitations of generic AI in specialized procedural domains.

A nuanced analysis of CA-GPT's performance revealed strengths and areas for further enhancement. While the system consistently achieved expert-level agreement on critical parameters such as minimum stent area (MSA) (100% agreement), its performance on metrics like pretreatment device selection (73.6%) and stent expansion assessment (78.4%) displayed inherent variability typical of interventional practice. Experienced operators often rely on subtle clinical judgment, such as accommodating complex lesion anatomies, which can occasionally lead to deviations from guideline-based thresholds. A similar issue was observed in AI-OCT studies. For example, the AutoOCT system performed well in standard cases, but struggled with complex lesions, such as heavily calcified or irregularly shaped lesions, where clinician decisions diverged from AI outputs.[32] This highlights that AI should act as an intelligent assistant, augmenting rather than replacing human decision-making. The CA-GPT-based AI-OCT system should be viewed similarly-while it excels in standardizing certain PCI decisions, clinical expertise remains paramount, especially in complex scenarios.[33-37]

The clinical implementation of the CA-GPT-based AI-OCT system addresses two major limitations in current OCT-guided PCI practice: operator-dependent interpretation variability and the time-intensive nature of manual analysis. Our system achieves expert-level agreement while dramatically reducing interpretation time from several minutes to mere seconds. This efficiency gain, combined with standardized decision outputs, positions CA-GPT as a valuable tool for enhancing workflow efficiency and procedural consistency. By providing immediate, evidence-based feedback on critical parameters such as stent expansion, apposition (93.2%), and edge dissection (97.7%), the system can reduce reliance on post-hoc analysis and minimize intra-procedural uncertainty, potentially accelerating the adoption of OCT guidance in routine practice.

From an educational perspective, the CA-GPT-based AI-OCT system represents a transformative platform for competency-based training in interventional cardiology. Junior operators can leverage the system's transparent reasoning pathways to understand the evidence base underlying specific recommendations, compare their decisions against guideline-concordant standards, and receive structured feedback on discrepancies. This educational dimension not only shortens the learning curve for OCT interpretation but also promotes consistent, high-quality practice across varying experience levels.[38] The explainable nature of the system's outputs, supported by its RAG framework, creates an optimal environment for developing both technical skills and clinical reasoning abilities.

Several limitations of this study must be acknowledged. (i) First, this was a single-center, retrospective study, limiting the generalizability of the findings to other populations and imaging systems. External validation across multiple centers and different clinical settings is required to confirm these results. (ii) The study did not include long-term clinical endpoint assessments such as MACE or mortality. However, this omission was intentional, as the primary aim of this work was to provide an initial, structured evaluation of CA-GPT's decision-making consistency rather than its long-term prognostic value. Future prospective studies will be required to determine whether this consistency translates into improved clinical outcomes. (iii) Furthermore, while this study demonstrates significant advancements in real-time decision support, future work should continue to explore the hybridization of human-AI workflows, where AI recommendations are dynamically weighted against operator judgment, to optimize clinical outcomes. These considerations will be essential for broader clinical implementation and integration of CA-GPT-based AI-OCT system into routine practice.

**Conclusion**

This study demonstrates that the domain-specific, CA-GPT-based AI-OCT system significantly outperforms both a general-purpose large language model (ChatGPT-5) and junior interventionalists in pre-PCI procedural planning, while also achieving superior overall agreement in post-PCI assessment. Its performance was particularly strong in complex clinical scenarios, such as lesions with significant ischemia and ACS presentations. By utilizing the CA-GPT to drive automated OCT analysis within a RAG framework, this system provides a standardized and reliable approach to intravascular imaging interpretation and decision support. These findings highlight the potential of domain-specific AI to augment operator expertise, bridge skill gaps, and enhance the precision and standardization of OCT-guided PCI. The promising performance of the CA-GPT-based AI-OCT system warrants further validation in large, multicenter, prospective studies.

## Supplementary materials

The supplementary materials used to describe the scoring criteria, physician characteristics, other details of the model and results.

## Acknowledgements
The authors are grateful to Vivolight Medtech (China) for providing OCT data interface support (P80) and technical training on the CA-GPT-based AI-OCT system.

## Declarations

### *Ethics approval and consent to participate*
This study was approved by the Institutional Ethics Committee of the Institution for National Drug Clinical Trials, Tangdu Hospital, Fourth Military Medical University (Ethics Approval No. K-HG-202511-15). Informed consent was waived for this analysis of anonymized clinical data.

### *Consent for publication*
Not applicable.

### *Data Availability*
All data generated or analysed during this study are included in this published article and its supplementary information files.

### *Competing interests*
The authors declare that they have no competing interests.

## Funding
This work was supported by National Key R&D Program of China (grant numbers 2025YFC2422703, 2025YFC2422703), Shenzhen Science and Technology Program (grant numbers KJZD20240903103304006), Health Science and Technology Innovation Capacity Improvement Program of Shaanxi Province (No. 2024TD-09), Key R&D Program of Shaanxi Province - Key Industrial Chain (No.2024SF-ZDCYL-01-03), Sanqin Talents Special Support Program (Yan Li).

## Authors' contributions
This study represents a clinician-engineer collaboration between the team of Director Yan Li at Tangdu Hospital of Fourth Military Medical University and that of Professor Rui Zhu at the Research Institute of Tsinghua University in Shenzhen. YL, WF, CW and WM conceived and designed the study. CW, WM, HL, JH, YC, MZ, JY, DZ, WG, BW, YX, WZ, ZW and YL formulated the PCI strategy for the senior expert group. ZL, PL, JZ, PZ, and CQ formulated the strategy for the young expert group. WF, CW, WM, and XN collected and analysed the data. RZ, YC, QL, and RM provided AI model information. WF wrote the first draft of the manuscript. YL edited the manuscript. WF, CW, and WM critically revised the manuscript. All the authors contributed to writing the paper and agreed with the manuscript results and conclusions. All authors read and approved the final manuscript.

**Table 1** Baseline Clinical and Imaging Characteristics of the Study Population

| Characteristic | N | Total (96 patients, 160 lesions) |
|---|---|---|
| Patients | 96 | |
| Age, years | | 60.2±10.9 |
| Male sex | | 66 (68.8%) |
| Height, cm | 91 | 166.5±7.5 |
| Weight, kg | 87 | 69.5±13.9 |
| BMI, kg/m2 | 86 | 24.9±4.0 |
| Smoking status | 88 | |
|     Previous | | 4 (4.55%) |
|     Current | | 42 (47.73%) |
|     Never | | 42 (47.73%) |
| Hypertension | 91 | 54 (59.3%) |
| Diabetes | 91 | 29 (31.9%) |
| Hypercholesterolemia | 94 | 6 (6.4%) |
| Previous MI | 94 | 4 (4.3%) |
| Previous PCI | 94 | 6 (6.4%) |
| Disease diagnosed | 93 | |
|     AMI | | 21 (22.6%) |
|     Unstable angina | | 56 (60.2%) |
|     SCAD | | 11 (11.8%) |
| PCI Strategy | | |
|     DES | | 77 (80.2%) |
|     DCB | | 13 (13.5%) |
|     No PCI performed | | 6 (6.3%) |
| Lesion distribution | 160 | |
|     LM-LAD | | 4 (2.5%) |
|     LAD | | 93 (58.1%) |
|     LCx | | 14 (8.8%) |
|     RCA | | 49 (30.6%) |
| Lesion length, mm | | 33.02±13.67 |
| Minimum lumen diameter | | 2.17±0.68 |
| Minimum lumen area, mm2 | | 3.94±2.42 |
| Calcium score | 72 | |
|     =4 | | 9 (12.5%) |
|     ＜4 | | 63 (87.5%) |
| Percentage diameter stenosis, % | | 29.4 (14.8-47.08) |
| Percentage area stenosis, % | | 49.55 (28-73.45) |
| MSA | 86 | 5.54±1.88 |
| MEI, % | 86 | 70.0±14.34 |
| Pre-PCI OCT-FFR | 72 | 0.74±0.08 |
| Post-PCI OCT-FFR | 88 | 0.89±0.06 |

BMI, Body mass index; MI, Myocardial infarction; PCI, Percutaneous coronary intervention; AMI, Acute myocardial infarction; SCAD, Stable coronary artery disease; DES, Drug-eluting stent; DCB, Drug-coated balloon; OCT, Optical coherence tomography; LM, Left main; LAD, Left anterior descending; LCx, Left circumflex; RCA,

Right coronary artery; MSA, Minimum stent area; MEI, Stent expansion index; OCT-FFR, Optical coherence tomography-fractional flow reserve

**Table 2** Comparison of decision agreement

| Outcome measure | CA-GPT | ChatGPT-5 | Junior physicians | P-value | P-value (CA-GPT vs. ChatGPT) | P-value (CA-GPT vs. Junior physicians) |
|---|---|---|---|---|---|---|
| **Pre-PCI agreement (0-5)** | 5 (3.75-5) | 3 (2-4) | 4 (3-4) | P<0.001 | P<0.001 | P<0.001 |
|   Revascularization judgement | 95.8% | 87.5% | 93.1% | **0.045** | 0.124 | 1.000 |
|   Pretreatment device type | 73.6% | 37.5% | 61.1% | P<0.001 | P<0.001 | 0.330 |
|   Pretreatment device sizing | 70.8% | 40.3% | 61.1% | P<0.001 | P<0.001 | 0.795 |
|   Stent diameter (±0.5mm) | 90.3% | 63.9% | 72.2% | P<0.001 | P<0.001 | **0.011** |
|   Stent length (±5mm) | 80.6% | 54.2% | 52.8% | P<0.001 | P<0.001 | **0.002** |
| **Post-PCI agreement (0-5)** | 5 (4.75-5) | 4 (4-5) | 5 (4-5) | P<0.001 | P<0.001 | **0.015** |
|   MSA | 100.0% | 100.0% | 95.5% | **0.018** | - | 0.268 |
|   Stent expansion | 78.4% | 33.0% | 84.1% | P<0.001 | P<0.001 | 1.000 |
|   Stent apposition | 93.2% | 88.6% | 76.1% | P<0.001 | 0.402 | **0.011** |
|   Severe dissection | 97.7% | 97.7% | 93.2% | **0.018** | - | 0.268 |
|   Significant tissue prolapse | 100.0% | 100.0% | 97.7% | 0.135 | - | - |

PCI, Percutaneous coronary intervention; MSA, minimum stent area; AI, Artificial intelligence; OCT, Optical coherence tomography

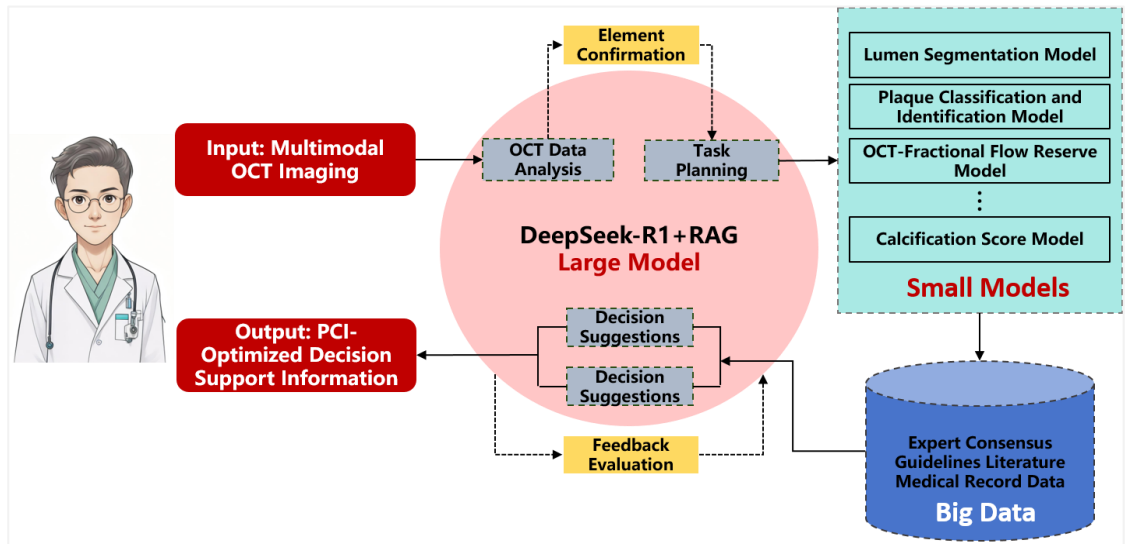

**Figure 1** CA-GPT-based AI-OCT System Decision Support Algorithm Framework

PCI, Percutaneous coronary intervention; OCT, Optical coherence tomography; RAG, Retrieval-augmented generation

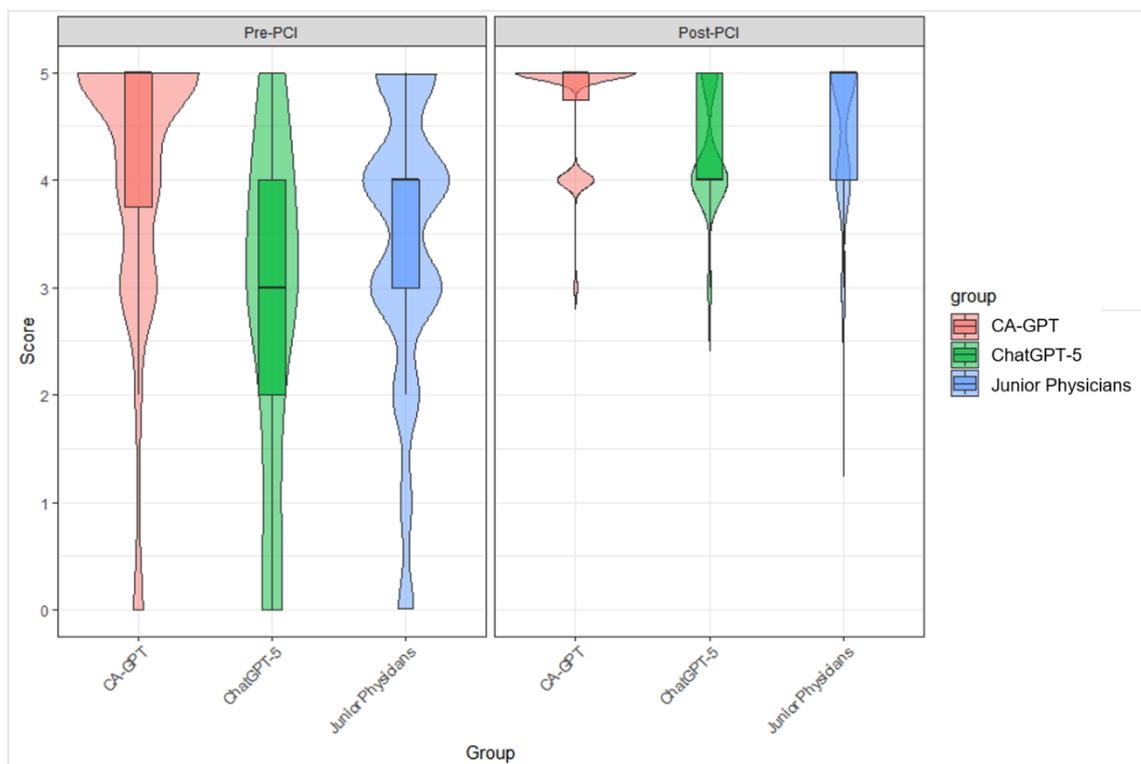

**Figure 2** Performance distribution of CA-GPT, ChatGPT-5 and Junior Physicians

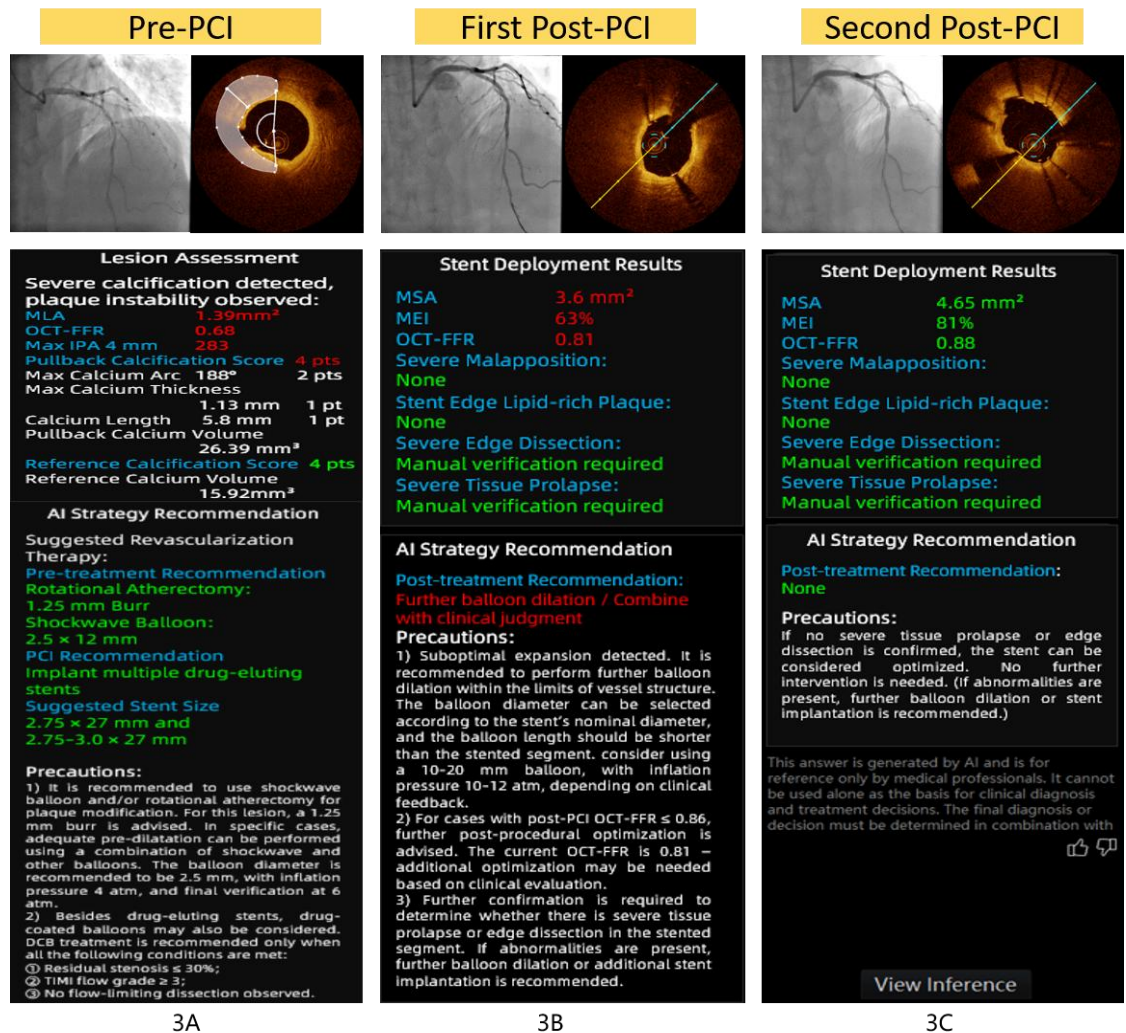

**Figure 3** Introduction to an CA-GPT-based AI-OCT System Assisted Decision-Making Case

The patient is a 61-year-old male with an initial diagnosis of coronary atherosclerotic heart disease and unstable angina. Figure 3A: Pre-procedural angiography and OCT imaging revealed a 50% stenosis at the ostium of the LAD, a 75% stenosis in the proximal to mid segment of the LAD with visible diffuse calcification, and a 50% stenosis in the distal LAD. CA-GPT recommended pre-treatment using rotational atherectomy (1.25 mm burr) / a Shockwave balloon (2.5 × 12 mm), and suggested PCI with implantation of multiple drug-eluting stents (stent size 2.75-3.0 × 27 mm). The actual procedure employed a 2.75×12 mm Shockwave balloon, which was deployed 6 times at the calcified lesion for pre-treatment. This was followed by implantation of a distal stent (2.75×38 mm) and a proximal stent (3.0×23 mm). Figure 3B: The first post-procedural angiography and OCT imaging showed underexpansion of the stent. CA-GPT recommended further balloon dilation. Accordingly, during the actual procedure, a 2.75×12 mm Shockwave balloon was advanced to the underexpanded segment of the stent and inflated 4 times at 6 atm. Figure 3C: The second post-procedural angiography and OCT imaging demonstrated good stent apposition and expansion, with no edge dissection or hematoma. Multi-angle angiography confirmed TIMI grade 3 flow in the LAD, and the procedure was concluded. Simultaneously, CA-GPT indicated favorable results, confirming the absence of edge dissection and tissue prolapse.

SUPPLEMENTARY MATERIALS

**COMPARE: Clinical Optimization with Modular Planning and Assessment via RAG-Enhanced AI-OCT: Superior Decision Support for Percutaneous Coronary Intervention Compared to ChatGPT-5 and Junior Operators**

**Table S1.** scoring criteria of different PCI strategies

| No. | Evaluation Metric | Type | Evaluation Criteria | Remarks |
|---|---|---|---|---|
| Pre-PCI | | | | |
| 1 | Revascularization | Qualitative | (1) Score 1 if consistent between both; (2) Score 0 if inconsistent | |
| 2 | Preparation Measures | Qualitative | (1) Score 1 if AI-suggested preparation measures include those actually used; (2) Score 0 if not | Mentions in the notes are also included |
| 3 | Preparation Device Size | Semi-quantitative | (1) Score 1 if preparation device diameter is within ±0.5 mm of actual use (if multiple sizes used, score 1 if AI-suggested size falls within actual range); (2) Score 0 if difference exceeds ±0.5 mm | In clinical practice, device selection and size may be adjusted dynamically based on preparation results, so multiple sizes might be used |
| 4 | Stent/Drug Balloon Diameter | Semi-quantitative | (1) Score 1 if diameter difference within ±0.5 mm; (2) Score 0 if difference exceeds ±0.5 mm | |
| 5 | Stent/Drug Balloon Length | Semi-quantitative | (1) Score 1 if length difference within ±5 mm; (2) Score 0 if difference exceeds ±5 mm | |
| Post-PCI | | | | |
| 6 | Minimum Stent Area | Quantitative | Score 1 if automatically calculated MSA is directionally consistent with recorded value (both >4.5 or both <4.5); Score 0 if inconsistent | |
| 7 | Stent Expansion | Semi-quantitative | Score 0 if CA-GPT strategy is inconsistent with actual strategy; Score 1 if consistent | Inconsistencies include: 1. CA-GPT suggests post-dilation due to under-expansion but actual records show good expansion; 2. CA-GPT doesn't suggest post-dilation but records show under-expansion and post-dilation was |

| No. | Evaluation Metric | Type | Evaluation Criteria | Remarks |
|---|---|---|---|---|
| | | | | performed; 3. If actual strategy doesn't mention expansion status but post-dilation was performed, it is considered under-expansion |
| 8 | Stent Apposition | Semi-quantitative | Score 0 if CA-GPT strategy is inconsistent with actual strategy; Score 1 if consistent | Inconsistencies include: 1. CA-GPT suggests post-dilation due to malapposition but actual records show good apposition; 2. CA-GPT doesn't suggest post-dilation but records show malapposition and post-dilation was performed; 3. If actual strategy doesn't mention apposition status but post-dilation was performed, it is considered malapposition |
| 9 | Severe Dissection | Semi-quantitative | Prerequisite: Severe dissection (angle ≥60° and length ≥3mm, extending to media) manually confirmed. Score 0 if CA-GPT strategy inconsistent with actual strategy; Score 1 if consistent | |
| 10 | Significant Tissue Prolapse | Semi-quantitative | Prerequisite: Significant tissue prolapse (prolapse area ≥10% of stent area) manually confirmed. Score 0 if CA-GPT strategy inconsistent with actual strategy; Score 1 if consistent | |

PCI, Percutaneous coronary intervention; OCT, Optical coherence tomography; MSA, Minimum stent area

**Table S2.** Characteristics of the participating physicians

| Characteristic | Junior physicians | Operating physicians |
| --- | --- | --- |
| Number | 4 | 14 |
| Highest academic degree | | |
|   Bachelor | 1 (25.0%) | 1 (7.1%) |
|   Master | 3 (75.0%) | 5 (35.7%) |
|   Doctor | 0 (0.0%) | 8 (57.1%) |
| Academic title | | |
|   Resident | 2 (50.0%) | 0 (0.0%) |
|   Attending | 1 (25.0%) | 0 (0.0%) |
|   Associate Chief | 1 (25.0%) | 9 |
|   Chief | 0 (0.0%) | 5 |
| Experience in PCI, years | | |
|   ≤5 | 4 (100.0%) | 0 (0.0%) |
|   5-10 | 0 (0.0%) | 2 (14.3%) |
|   10-15 | 0 (0.0%) | 8 (57.1%) |
|   >15 | 0 (0.0%) | 4 (28.6%) |
| Annual PCI volume, n | | |
|   ≤100 | 2 (50.0%) | 0 (0.0%) |
|   100-250 | 2 (50.0%) | 5 (35.7%) |
|   250-350 | 0 (0.0%) | 5 (35.7%) |
|   >350 | 0 (0.0%) | 4 (28.6%) |
| OCT-guided PCI experience, years | | |
|   ≤5 | 4 (100%) | 0 (0.0%) |
|   >5 | 0 (0.0%) | 14 (100%) |
| OCT-guided PCI cases per year, n | | |
|   ≤80 | 4 (100%) | 5 (35.7%) |
|   80-120 | 0 (0.0%) | 6 (42.9%) |
|   120-160 | 0 (0.0%) | 3 (21.4%) |

The OCT experience of junior physicians was calculated based on their role as assistants. PCI, Percutaneous coronary intervention

**Table S3.** Subgroup analysis of decision agreement

| Subgroup | Lesions, n | CA-GPT | ChatGPT-5 | Junior physicians | P-value | P-value (CA-GPT vs. ChatGPT) | P-value (CA-GPT vs. Junior physicians) |
|---|---|---|---|---|---|---|---|
| **Pre-PCI** | | | | | | | |
| Lesion distribution | | | | | | | |
|   LAD | 46 | 4 (3.25-5) | 3 (2-3.75) | 3.5 (3-4) | P＜0.001 | P＜0.001 | 0.076 |
|   LCx/RCA | 25 | 5 (4-5) | 4 (1-4) | 4 (2-4) | **0.002** | **0.012** | **0.005** |
| Pre-PCI OCT-FFR | | | | | | | |
|   OCT-FFR ≤ 0.80 | 57 | 5 (4-5) | 3 (3-4) | 4 (3-4) | P＜0.001 | P＜0.001 | P＜0.001 |
|   OCT-FFR＞0.80 | 15 | 5 (3.5-5) | 0 (0-3.5) | 5 (2-5) | **0.002** | **0.031** | 1.000 |
| Disease diagnosed | | | | | | | |
|   ACS | 58 | 5 (3.25-5) | 3 (2-4) | 4 (3-4.75) | P＜0.001 | P＜0.001 | **0.005** |
|   SCAD | 7 | 4 (4-4.5) | 3 (2.5-3.5) | 4 (3-4) | **0.029** | 0.143 | 1.000 |
| Calcification Severity | | | | | | | |
|   Mild (calcium score＜4) | 63 | 5 (4-5) | 3 (2-4) | 4 (3-4) | P＜0.001 | P＜0.001 | P＜0.001 |
|   Severe (calcium score=4) | 9 | 5 (3-5) | 3 (1-3) | 3 (2-4) | **0.040** | 0.105 | 0.708 |
| **Post-PCI** | | | | | | | |
| Lesion distribution | | | | | | | |
|   LAD | 47 | 5 (4-5) | 4 (4-4) | 5 (4-5) | P＜0.001 | P＜0.001 | 0.408 |
|   LCx/RCA | 38 | 5 (5-5) | 4 (4-5) | 5 (4-5) | P＜0.001 | P＜0.001 | **0.015** |
| Disease diagnosed | | | | | | | |
|   ACS | 71 | 5 (4.5-5) | 4 (4-5) | 5 (4-5) | P＜0.001 | P＜0.001 | 0.060 |
|   SCAD | 10 | 5 (5-5) | 4 (4-4) | 4.5 (4-5) | **0.016** | **0.032** | 0.267 |

PCI, Percutaneous coronary intervention; AI, Artificial intelligence; OCT, Optical coherence tomography; OCT-FFR: Optical coherence tomography- fractional flow reserve; LAD, Left anterior descending; LCx, Left circumflex; RCA: Right coronary artery; ACS: Acute coronary syndrome; SCAD, Stable coronary artery disease

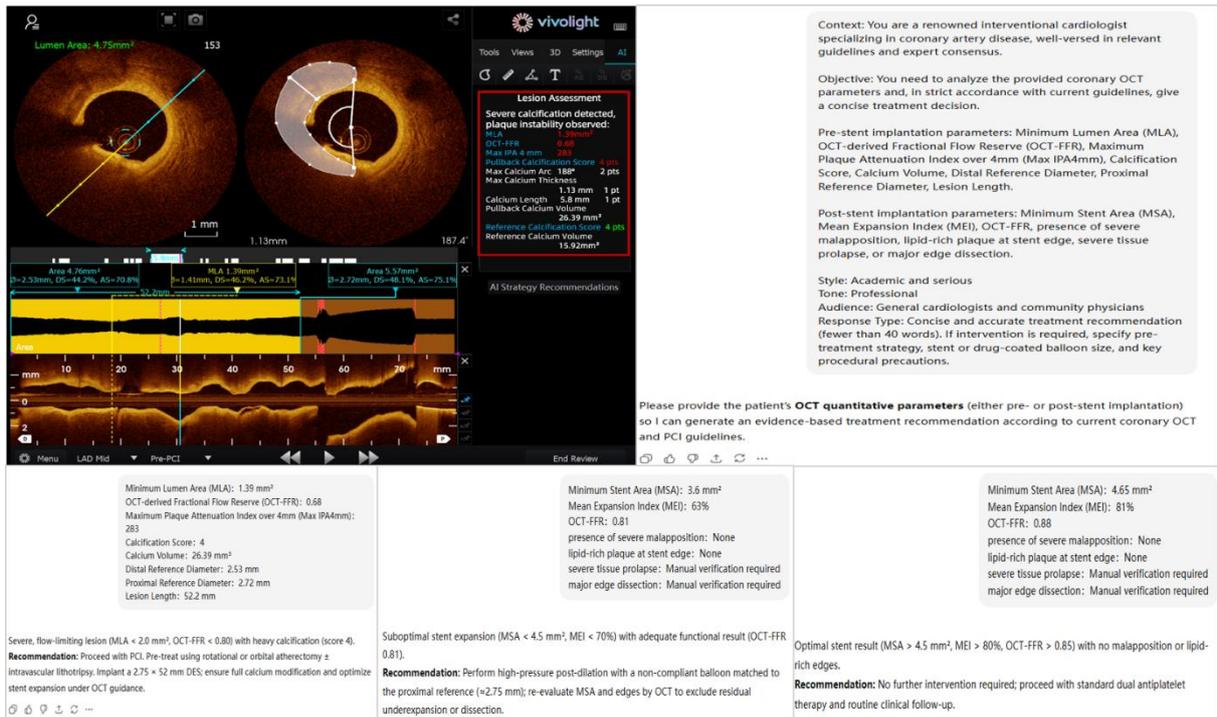

**Figure S1.** Input and Output of ChatGPT-5

MLA, Minimum lumen area; OCT-FFR: Optical coherence tomography- fractional flow reserve; IPA, Index of plaque attenuation; DS, Diameter stenosis; AS, Area stenosis; MEI, Stent expansion index; MSA, Minimum stent area

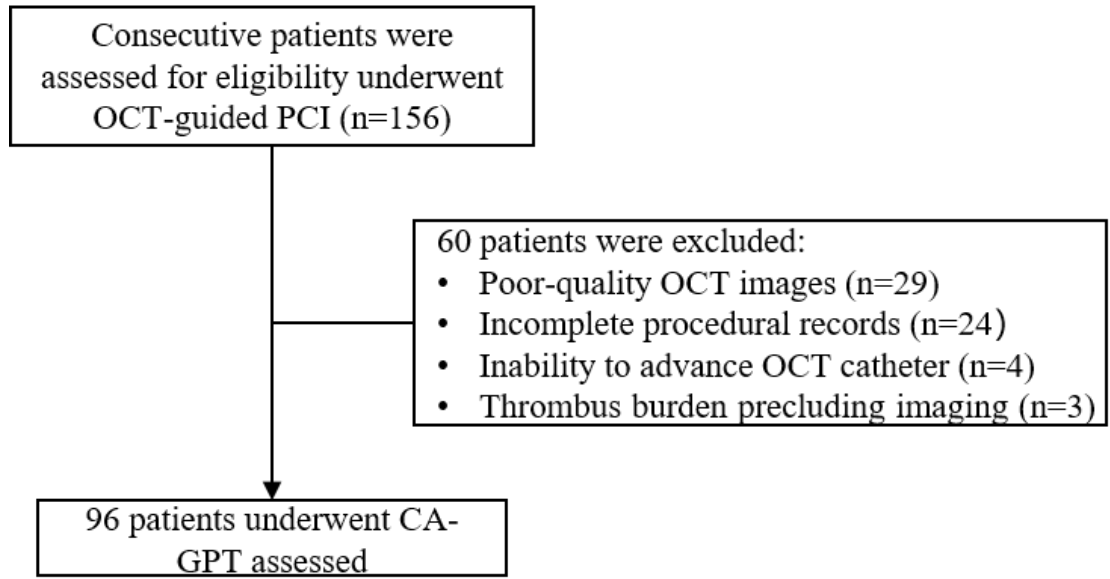

**Figure S2.** Patient Selection Flowchart

PCI, Percutaneous coronary intervention; OCT, Optical coherence tomography

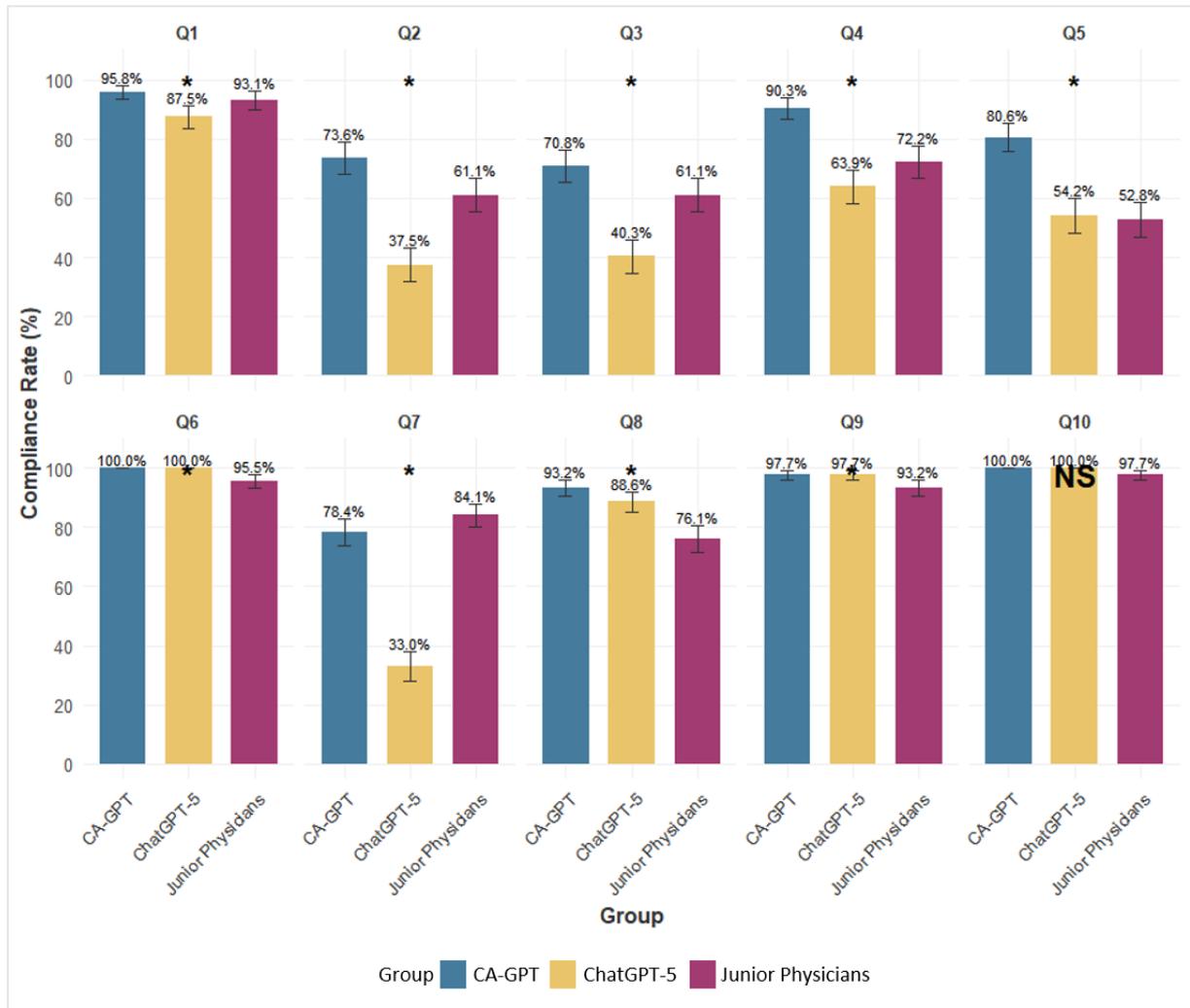

**Figure S3.** Performance distribution across scoring items

* Compliance rates show a significant difference; NS, Not significant